\documentclass[letterpaper, 10 pt, conference]{ieeeconf}  

\IEEEoverridecommandlockouts                              

\usepackage{amsmath,amsfonts}
\usepackage{algorithmic}
\usepackage{algorithm}
\usepackage{array}
\usepackage[caption=false,font=normalsize,labelfont=sf,textfont=sf]{subfig}
\usepackage{textcomp}
\usepackage{stfloats}
\usepackage{url}
\usepackage{verbatim}
\usepackage{graphicx}
\usepackage{cite}

\usepackage{color,xcolor}
\usepackage{balance}
\usepackage[misc]{ifsym}
\usepackage{multirow}
\usepackage{booktabs}
\usepackage{hyperref}
\usepackage{makecell}

\usepackage{tikz,xcolor,hyperref}
\definecolor{lime}{HTML}{A6CE39}
\DeclareRobustCommand{\orcidicon}{%
    \begin{tikzpicture}
    \draw[lime, fill=lime] (0,0)
    circle [radius=0.16]
    node[white] {{\fontfamily{qag}\selectfont \tiny ID}};    \draw[white, fill=white] (-0.0625,0.095)
    circle [radius=0.007];    \end{tikzpicture}
    \hspace{-2mm}}
\foreach \x in {A, ..., Z}{%
    \expandafter\xdef\csname orcid\x\endcsname{\noexpand\href{https://orcid.org/\csname orcidauthor\x\endcsname}{\noexpand\orcidicon}}
    }

\title{\LARGE \bf
GSDC Transformer: An Efficient and Effective Cue Fusion for Monocular Multi-Frame Depth Estimation}

\author{Naiyu Fang\orcidA{}~\IEEEmembership{Student Member,~IEEE}, Lemiao Qiu\orcidB{}~\IEEEmembership{Member,~IEEE}, Shuyou Zhang, \\Zili Wang\orcidD{}~\IEEEmembership{Member,~IEEE}, Zheyuan Zhou\orcidF{}, and Kerui Hu\orcidE{}
\thanks{This work has been submitted to the IEEE for possible publication. Copyright may be transferred without notice, after which this version may no longer be accessible.}
\thanks{This work was supported by the National Natural Science Foundation of China (No.52375271); the Natural Science Foundation of Zhejiang Province (No.LY23E050011).}%
\thanks{The authors are with State Key Laboratory of Fluid Power \& Mechatronic Systems, Zhejiang University, Hangzhou, 310027, China (e-mail: FangNaiyu@zju.edu.cn; qiulm@zju.edu.cn; zsy@zju.edu.cn; ziliwang@zju.edu.cn; zheyuanzhou@zju.edu.cn; hkr457@zju.edu.cn) \it(Corresponding authors: Lemiao Qiu)}%
}

\begin{document}

\maketitle
\thispagestyle{empty}
\pagestyle{empty}

\begin{abstract}
Depth estimation provides an alternative approach for perceiving 3D information in autonomous driving. Monocular depth estimation, whether with single-frame or multi-frame inputs, has achieved significant success by learning various types of cues and specializing in either static or dynamic scenes. Recently, these cues fusion becomes an attractive topic, aiming to enable the combined cues to perform well in both types of scenes. However, adaptive cue fusion relies on attention mechanisms, where the quadratic complexity limits the granularity of cue representation. Additionally, explicit cue fusion depends on precise segmentation, which imposes a heavy burden on mask prediction. To address these issues, we propose the GSDC Transformer, an efficient and effective component for cue fusion in monocular multi-frame depth estimation. We utilize deformable attention to learn cue relationships at a fine scale, while sparse attention reduces computational requirements when granularity increases. To compensate for the precision drop in dynamic scenes, we represent scene attributes in the form of super tokens without relying on precise shapes. Within each super token attributed to dynamic scenes, we gather its relevant cues and learn local dense relationships to enhance cue fusion. Our method achieves state-of-the-art performance on the KITTI dataset with efficient fusion speed.
\end{abstract}

\section{Introduction}
\label{sec1}

\begin{figure}[!ht]
\centering
\includegraphics[scale=.35]{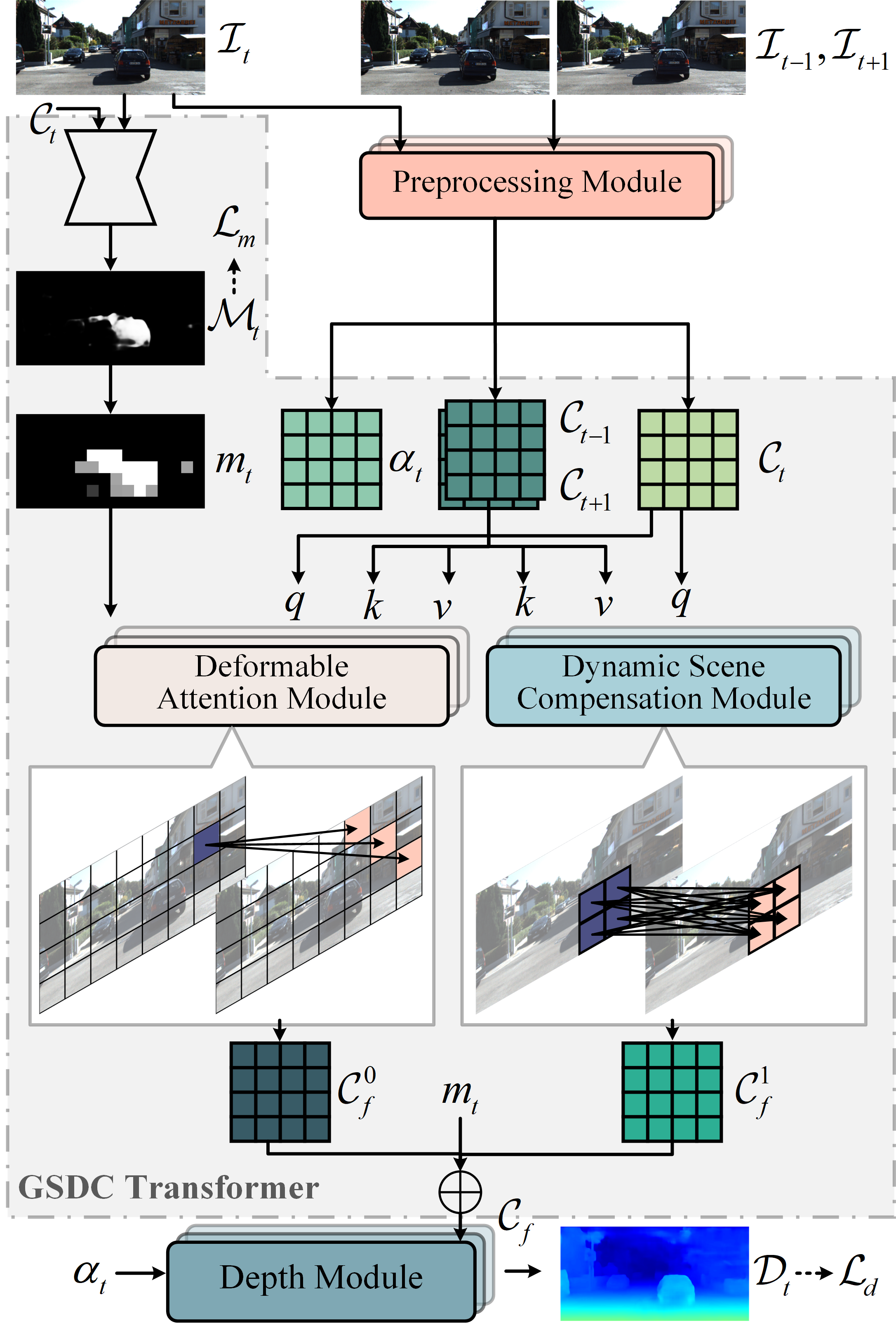}
\caption{The framework of monocular multi-frame depth estimation. The main focus of this paper is the cue fusion process, which is addressed by introducing a novel GSDC Transformer denoted by the gray block. GSDC Transformer is conditioned on the cost volumes of the target frame and adjacent frames, and yields a fused cost volume. It aims to increase the interaction granularity in a computation-saving manner and compensate for the precision drop in dynamic scenes.}
\label{fig1}
\end{figure}

As a step toward visual-only autonomous driving, depth estimation plays a crucial role in providing additional distance information for 3D object detection {\color{blue}\cite{zhou2022sgm3d,su2023opa,mouawad2022time}} and scene understanding {\color{blue}\cite{humblot2022navigation,can2022understanding}}, potentially replacing the need for LiDAR sensors. Both single-frame {\color{blue}\cite{laina2016deeper,liu2019neural,jiao2018look}} and multi-frame {\color{blue}\cite{mi2022generalized,ranftl2016dense,klingner2020self}} monocular depth estimations have achieved remarkable success by leveraging various cues {\color{blue}\cite{bae2022multi}}. However, their estimation mechanisms have limitations in either static or dynamic scenes {\color{blue}\cite{li2023learning}}. To circumvent their limitation, how to fuse these cues {\color{blue}\cite{li2023learning,bae2022multi}} is blooming in the community. In this paper, we focus on this cue fusion topic, to chase an efficient and effective component that enhances monocular multi-frame depth estimation.

Monocular depth estimation provides a low-hardware-cost scheme to perceive distance. The single-frame approach learns spatial context cues like texture gradient and utilizes a deep feature extractor to estimate pixel-level depth {\color{blue}\cite{bae2022multi}}. Although significant improvements have been made through model and mechanism upgrades {\color{blue}\cite{fu2018deep,bhat2021adabins}}, it remains an ill-posed problem, leading to artifacts in static scenes. To tackle it, inspired by stereo match {\color{blue}\cite{gu2020cascade}}, the multi-frame approach exploits cost volume {\color{blue}\cite{newcombe2011dtam}} to match features between adjacent frame at different depth hypotheses. This enables significant progress in the estimation precision at static scenes by leveraging geometric cues. However, it encounters challenges in dynamic scenes due to large displacements that violate geometric consistency. There are two prevailing methods to mitigate the failure in dynamic scenes. The first method involves explicitly distinguishing dynamic scenes at the pixel level using masks {\color{blue}\cite{wimbauer2021monorec,feng2022disentangling}} or optical flow {\color{blue}\cite{luo2020consistent}}. It replaces the estimated results or cost volume in dynamic scenes of the multi-frame approach with those from the single-frame approach and fine-tunes them at the loss level. With the emergence of Vision Transformer {\color{blue}\cite{dosovitskiy2021an}}, the second method focuses on learning the spatial relationship between texture cues and geometric cues using attention mechanisms. It generates a fused result that adaptively emphasizes cues in scenes without relying on an explicit mask. Intrinsically, both methods aim to fuse cues to improve depth estimation performance in both static and dynamic scenes.

However, there are still some issues with the aforementioned fusion methods. 1) Due to the low pixel proportion of dynamic scenes, the detection and segmentation of pixel-level dynamic scenes require high shape accuracy, which poses a significant burden on the mask prediction module. Incorrect shapes can lead to erroneous replacements; 2) Due to the inherent quadric complexity, Transformer-based adaptive cue fusion inevitably reduces the granularity of cue representations through downsampling, impacting the final estimation precision at the pixel level, while full attention also affects fusion efficiency.

To tackle these issues, we propose a novel Globally Sparse and Dense-Compensated Transformer (GSDC Transformer), which functions as a crucial component for cue fusion in monocular multi-frame depth estimation. We strive to establish a learning paradigm for efficient and effective cue fusion, where the core idea is to fine overall granularity and compensate for local dynamic scenes. The contribution of this paper is three-fold:

(1) We propose a super dynamic scenes mask to alleviate the need for precise segmentation. It represents quantal scene attributions in the form of super token and provides enough candidates for compensating dynamic scenes, while its prediction relies on a lightweight CNN model, thus ensuring efficient fusion processes.

(2) We propose to enhance the overall fusion granularity and compensate for dynamic scenes. The deformable attention module learns fine-scale spatial relationships between cues. To mitigate the precision drop caused by sparse relationships in dynamic scenes, a dynamic compensation module supplements local dense relationships for each super token.

(3) Experiments demonstrate that our method achieves state-of-the-art performance on the KITTI dataset {\color{blue}\cite{geiger2012we}}, ranking first and second for overall and dynamic scenes, respectively. Furthermore, our fusion method achieves nearly 20\% savings in FLOPs compared to {\color{blue}\cite{li2023learning}}.

\section{Related Work}
\label{sec2}

\subsection{Single-Frame Depth Estimation}
\label{sec2.1}

Learning-based single-frame depth estimation has emerged with the advent of CNNs, utilizing an encoder-decoder structure to map RGB features to depth values at the pixel level. Laina \textit{et al}. {\color{blue}\cite{laina2016deeper}} proposed an end-to-end model for this task, eliminating the need for hand-crafted features and post-processing techniques. Due to the long-tailed distribution of depth values, direct regression of depth values encounters issues of slow convergence and local minima in end-to-end models. Several studies aimed to develop effective prediction heads and objective functions. Fu \textit{et al}. {\color{blue}\cite{fu2018deep}} employed the SID strategy to convert continuous depth into discrete values. Liu \textit{et al}. {\color{blue}\cite{liu2019neural}} regressed a depth probability distribution to construct a 3D depth probability volume. Jiao \textit{et al}. {\color{blue}\cite{jiao2018look}} utilized attention-driven loss to establish a connection between semantic segmentation and depth estimation. Yin \textit{et al}. {\color{blue}\cite{yin2019enforcing}} utilized geometric constraints in 3D space to penalize virtual normal directions.

Building on these frameworks and paradigms, some studies aimed to capture more global information in depth estimation. They achieved this by utilizing progressive upsampling to expand the receptive field or employing dilated convolutions {\color{blue}\cite{fu2018deep}}. With the emergence of Vision Transformers, Ranftl \textit{et al}. {\color{blue}\cite{ranftl2021vision}} applied this task within a pure attention framework and gained a substantial precision improvement. Bhat \textit{et al}. {\color{blue}\cite{bhat2021adabins}} further introduced global processing information and adaptively learned the range of discrete depth values based on specific instances.

\subsection{Multi-Frame Depth Estimation}
\label{sec2.2}
A traditional approach for multi-frame depth estimation involves capturing features across frames using a temporal model. Wang  \textit{et al}. {\color{blue}\cite{wang2019recurrent}} utilized a recurrent neural network, Patil \textit{et al}. {\color{blue}\cite{patil2020don}} employed a ConvLSTM model to learn spatiotemporal information, and Wang \textit{et al}. {\color{blue}\cite{wang2022itermvs}} utilized a GRU-based estimator to match feature similarity in the hidden state. Drawing inspiration from multi-view stereo {\color{blue}\cite{cheng2020deep}}, mainstream methods construct cost volumes {\color{blue}\cite{mi2022generalized}} to explore the 3D spatial relationships between multiple frames. This approach has achieved significant success in handling the ambiguity of monocular depth estimation, particularly for static scenes.

Due to the presence of moving objects in autonomous driving, such as vehicles, motorbikes, and pedestrians, and these dynamic scenes deviate from geometric assumptions, some studies endeavored to address the artifact of dynamic scenes in depth estimation. They partition dynamic scenes in calculating photometric loss {\color{blue}\cite{ranftl2016dense,klingner2020self}} and constructing cost volumes {\color{blue}\cite{watson2021temporal,feng2022disentangling}}, or model moving objects by leveraging the fundamental difference between inverse and forward projection {\color{blue}\cite{lee2021learning}}.

The multi-view stereo also offers a paradigm for unsupervised and self-supervised learning of multi-frame depth estimation. Geometry consistency {\color{blue}\cite{bian2019unsupervised}} serves as a fundamental component, supervising depth estimation through the projection and warping between frames. At the loss level, this constraint is further enhanced by introducing Depth Hints {\color{blue}\cite{watson2019self}} and employing a minimum reprojection loss {\color{blue}\cite{godard2019digging}}. At the mechanism level, Shu \textit{et al}. {\color{blue}\cite{shu2020feature}} proposed a feature-level projection, while Johnston \textit{et al}. {\color{blue}\cite{johnston2020self}} employed self-attention to capture features and constrained the discrete disparity volume.

\section{Methodology}
\label{sec3}

\subsection{Outline}
\label{sec3.1}
Monocular multi-frame depth estimation predicts a depth map ${{{\cal D}_t}}$ from the target frame ${{{\cal I}_t}}$ at time ${t}$ by learning information from adjacent frames ${{{\cal I}_{t - 1}},{{\cal I}_{t + 1}}}$. This paper focuses on information learning process between cues in the target frame and adjacent frames, i.e. cue fusion. As {\color{blue}Fig. \ref{fig1}} shows, in the preprocessing module, we exploit the depth converting method {\color{blue}\cite{li2023learning}} to obtain the pseudo cost volume ${{{\cal C}_t} \in {{ \mathbb{R} }^{H \times W \times D}}}$ of ${{{\cal I}_t}}$, where ${D}$ represents the number of discrete depth channels. We utilize SSIM-based photometric error method {\color{blue}\cite{wimbauer2021monorec}} to construct cost volumes ${{{\cal C}_{t - 1}},{{\cal C}_{t + 1}} \in {{ \mathbb{R} }^{H \times W \times D}}}$ conditioned on ${{{\cal I}_{t - 1}},{{\cal I}_t},{{\cal I}_{t + 1}}}$, and these cost volumes are concatenated at depth channel instead of weighted summation. Additionally, we employ a backbone (ResNet18 {\color{blue}\cite{he2016deep}} or Efficient-B5 {\color{blue}\cite{tan2019efficientnet}}) to extract the multi-scale features ${{\alpha _t}}$ from ${{{\cal I}_t}}$. Based on these inputs, we propose a novel GSDC Transformer to combine cues from adjacent frame into the target frame, resulting in the prediction of a fused cost volume ${{{\cal C}_f} \in {{ \mathbb{R} }^{H \times W \times D}}}$. Eventually, ${{{\cal C}_f}}$ is fed into a depth module {\color{blue}\cite{li2023learning}} to predict the depth map ${{{\cal D}_t} \in {{ \mathbb{R} }^{H \times W \times 1}}}$ of the target frame.

To achieve efficient and effective multi-frame fusion, we devise GSDC Transformer to consist of a deformable attention module, a super dynamic scene mask prediction module, and a dynamic scene compensation module. Inspired by Deformable DETR {\color{blue}\cite{zhu2021deformable}}, the deformable attention module (in {\color{blue} Sec. \ref{sec3.2.1}}) establishes sparse attention relationships between ${{{\cal C}_t}}$ and ${{{\cal C}_{t - 1}},{{\cal C}_{t + 1}}}$ for each token. By employing linear complexity instead of quadratic complexity, the deformable attention module can be implemented at a finer grain, enhancing the precision of overall scenes. To compensate for the precision drop in dynamic scenes, we adopt a lightweight CNN module to partition dynamic scenes (in {\color{blue} Sec. \ref{sec3.2.2}}) into super tokens without relying on precise shape prediction. Finally, the dynamic scene compensation module (in {\color{blue} Sec. \ref{sec3.2.3}}) focuses on learning local and dense attention relationships exclusively within dynamic scenes to facilitate its cue fusion.

\subsection{GSDC Transformer}
\label{sec3.2}

\subsubsection{Fine-Grained and Sparse Attention but Precision Drop in Dynamic Scenes}
\label{sec3.2.1}

According to {\color{blue}\cite{bae2022multi}}, monocular single-frame depth estimation relies on texture spatial context cues such as texture gradient, while monocular multi-frame depth estimation focuses on learning geometric cues between adjacent frames. Intrinsically, we aim to fuse the geometric cues from ${{{\cal C}_{t - 1}},{{\cal C}_{t + 1}}}$ with spatial context cues in ${{{\cal C}_t}}$ by learning their correspondence relationship, incorporating both types of cues in the fused cost volume ${{{\cal C}_f}}$. A previous study {\color{blue}\cite{li2023learning}} employed full attention to learn the relationships between these cues. However, the quadratic complexity of attention led to downsampling the cost volumes by a factor of ${4 \times }$ to manage computational costs. Unfortunately, this downsampling results in a coarse-grained token representation, leading to a precision drop in cue fusion.

\begin{figure*}[!h]
\centering
\includegraphics[scale=.36]{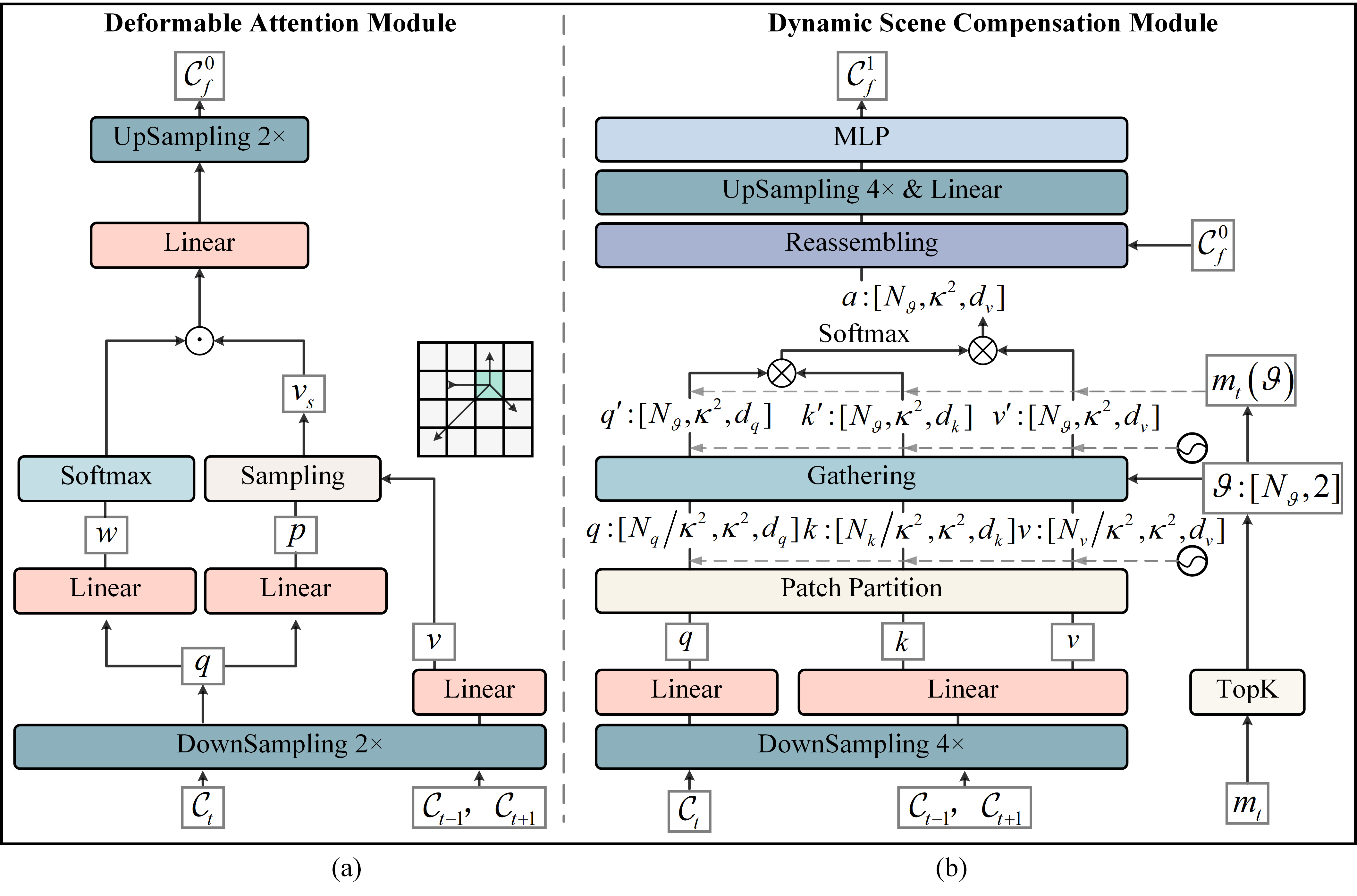}
\caption{The detailed structures of the deformable attention module and the dynamic scene compensation module. (a) The deformable attention module increases the granularity of cue representation with only downsampling ${2 \times }$ and employs sparse attention to save computation costs. (b) The dynamic scene compensation module gathers tokens attributed to super dynamic scene tokens and learns their local dense relationships to compensate for the precision drop.}
\label{fig2}
\end{figure*}

We aim to increase the granularity of cue representation while maintaining computational efficiency. To achieve this, a straightforward and well-known approach is to employ sparse attention instead of full attention. Since cost volumes are constructed in a pixel-aligned manner and most pixels exhibit minimal movement between adjacent frames, there is an opportunity to explore the correspondence relationship in the cue fusion via a sparse manner. Therefore, in the deformable attention module, we increase the overall cue granularity by downsampling tokens by a factor of ${2 \times }$, and the attention is sparse to ensure an efficient cue fusion. Experimental results presented in {\color{blue} Table \ref{table2}} demonstrate that sparse attention does not significantly compromise overall precision. Specifically, following the paradigm of deformable DETR {\color{blue}\cite{zhu2021deformable}}, as shown in {\color{blue}Fig. \ref{fig2}a}, we utilize ${{{\cal C}_t}}$ as the query to predict weights ${w \in {{ \mathbb{R} }^{{N_q} \times {N_s} \times 1}}}$ and offsets ${p \in {{ \mathbb{R} }^{{N_q} \times {N_s} \times 1}}}$, where ${{N_s}}$ represents the number of sample points with ${{N_s} \ll {N_v}}$. The concatenated ${{{\cal C}_{t - 1}},{{\cal C}_{t + 1}}}$ serve as the value, and ${{N_s}}$ points are sampled for each token, resulting in ${{v_s} \in {{ \mathbb{R} }^{{N_q} \times {N_s} \times {d_v}}}}$. Finally, the product of ${{v_s}}$ and ${w}$ yields the predicted fused cost volume ${{\cal C}_f^0}$ through projection and upsampling.

However, it is crucial to consider dynamic scenes despite their low pixel percentage {\color{blue}\cite{bae2022multi}}. Due to the noticeable positional shifts of moving objects between adjacent frames, more correspondence relationships need to be learned in cue fusion by maintaining the preservation and enhancement of spatial context cues. Unfortunately, as shown in {\color{blue} Table \ref{table2}}, the sparse attention fails to effectively handle dynamic scenes, leading to a significant precision drop.

\subsubsection{Super Dynamic Scene Mask}
\label{sec3.2.2}

\begin{figure}[!ht]
\centering
\includegraphics[scale=.30]{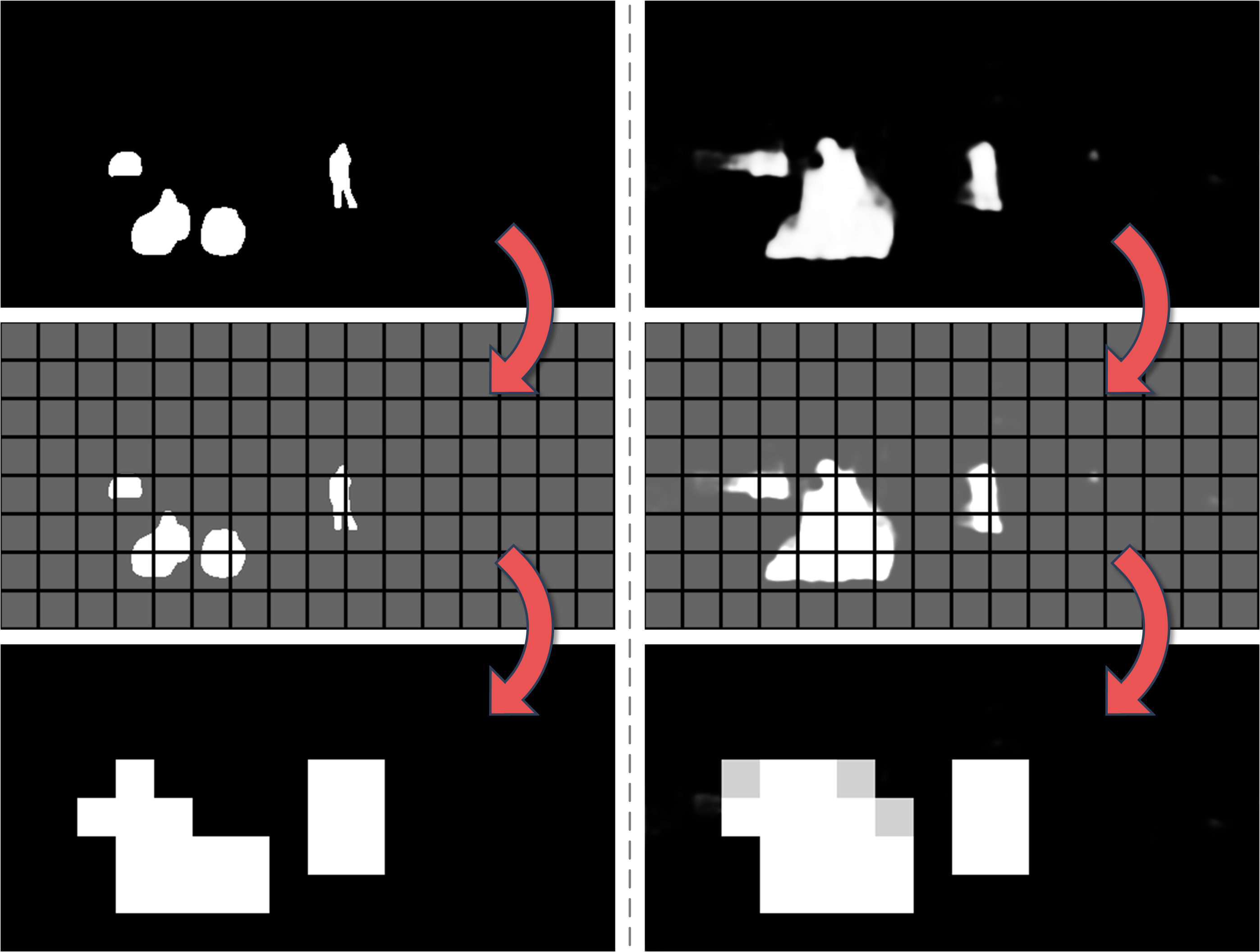}
\caption{The left and right parts represent the super dynamic scene masks of the predicted result and the ground truth, respectively. Compared to the original form, the super token form can represent dynamic scenes without relying on the precise shape.}
\label{fig3}
\end{figure}

To maintain our desired efficiency, a straightforward approach is to supplement full attention only to dynamic scenes. However, this approach heavily relies on an explicit semantic segmentation of dynamic scenes, necessitating a precise dynamic scene mask. In this section, our goal is to mitigate the reliance on this precise shape prior. Inspired by the window local attention mechanism {\color{blue}\cite{liu2021swin}}, as shown in {\color{blue}Fig. \ref{fig3}}, we exploit windows to partition the dynamic scene mask ${{{\cal M}_t}}$ into a super dynamic scene mask ${{m_t}}$, where ${\kappa  \times \kappa }$ tokens are grouped within each window to form a super token. Each super token is assigned a unified scene attribution as {\color{blue} Equ. (\ref{eq1})}.

\begin{equation}
\label{eq1}
\small
{m_t}\left( {x,y} \right) = \max \left\{ {{{\cal M}_t}\left( {u,v} \right)|u \in \left[ {x\kappa ,x\kappa  + \kappa } \right.),v \in \left[ {y\kappa ,y\kappa  + \kappa } \right.)} \right\}
\end{equation}

The super dynamic scene mask is designed to provide a quantal attribution for each super token, rather than representing the shape at the pixel level. This converting approach also reduces the prediction burden of the model. Specifically, we concatenate the target frame ${{{\cal I}_t}}$ and the pseudo cost volume ${{{\cal C}_t}}$, fed them into a lightweight U-Net {\color{blue}\cite{ronneberger2015u}} to predict ${{{\cal M}_t}}$, and convert it into ${{m_t}}$. As shown in {\color{blue}Fig. \ref{fig3}}, although the predicted dynamic scene mask may exhibit shape differences compared to the ground truth, it still provides an adequate number of super token candidates attributed to dynamic scenes.

\subsubsection{Compensate for Dynamic Scenes}
\label{sec3.2.3}

The super dynamic scene mask enables us to establish connections between each super token attributed to dynamic scenes and a local window in the Swin Transformer {\color{blue}\cite{liu2021swin}}, allowing for full attention among tokens within each super token. To achieve this, we propose a dynamic compensation module as shown in {\color{blue}Fig. \ref{fig2}b}, which takes ${{{\cal C}_t}}$ as the query, and ${{{\cal C}_{t - 1}},{{\cal C}_{t + 1}}}$ as the key and value, downsampling them by a factor of ${4 \times }$. We partition ${q,k,v}$ using windows of size ${\kappa  \times \kappa}$, and embed a learned position matrix at ${{{{N_q}} \mathord{\left/{\vphantom {{{N_q}} {{\kappa ^2}}}} \right.\kern-\nulldelimiterspace} {{\kappa ^2}}}}$ dimension.

Since the number of super tokens attributed to dynamic scenes varies across instances, and the standard Swin Transformer performs calculations in windows-parallel, it is important to maintain stability during training and inference. To address this, we employ a TopK function to extract ${{N_\vartheta }}$ (${{N_\vartheta } \ll \left( {{{HW} \mathord{\left/{\vphantom {{HW} {{\kappa ^2}}}} \right.\kern-\nulldelimiterspace} {{\kappa ^2}}}} \right)}$) candidate as {\color{blue}Equ. (\ref{eq2})}, where ${\vartheta }$ is the index matrix. Based on ${\vartheta }$, ${q',k',v'}$ are gathered from ${q,k,v}$ with a relative position embedding within each window.

\begin{equation}
\label{eq2}
{m_t}\left( \vartheta  \right),\vartheta  = TopK\left( {{m_t}} \right)
\end{equation}

In some cases, the tail parts of candidate tokens may have small scene values, indicating that they do not correspond to real dynamic scenes. To address this, we multiply ${q',k',v'}$ with ${{m_t}\left( \vartheta  \right)}$ to encourage the model to focus on real dynamic scene parts. Subsequently, we learn ${a \in {{ \mathbb{R} }^{{N_\vartheta } \times {\kappa ^2} \times {d_v}}}}$ using the well-known local window attention mechanism. In order to restore the original spatial structure, we reassemble ${a}$ using the base ${{\cal C}_f^0}$ and the index ${\vartheta }$. Through upsampling and MLP, we predict another fused cost volume ${{\cal C}_f^1}$.

Finally, we fuse ${{\cal C}_f^0}$ and ${{\cal C}_f^1}$ using weighted summation and a conv ${1 \times 1}$ layer as {\color{blue} Equ. (\ref{eq3})}.

\begin{equation}
\label{eq3}
{{\cal C}_f}{\rm{ = Conv}}\left[ {{\cal C}_f^0 \cdot \left( {1 - {m_t}} \right) + {\cal C}_f^1 \cdot {m_t}} \right]
\end{equation}

\subsubsection{Objective Function}
\label{sec3.2.4}

The objective function of super dynamic mask prediction module is formulated as {\color{blue} Equ. (\ref{eq4})} where ${{\rm{BCE}}\left(  \cdot  \right)}$ is the binary entropy loss, and ${{\tilde {\cal M}_t}}$ is the ground truth. The objective function of the depth module remains the same as {\color{blue}\cite{li2023learning}}, including the scale-invariant loss {\color{blue}\cite{bhat2021adabins}}, virtual normal loss {\color{blue}\cite{yin2021virtual}}, and a monocular depth loss.

\begin{equation}
\label{eq4}
{\ell _m}{\rm{ = BCE}}\left( {{{\cal M}_t},{{\tilde {\cal M}}_t}} \right)
\end{equation}

\section{Experiment}
\label{sec4}

\subsection{Implementation Details}
\label{sec4.1}

\subsubsection{Dataset}
\label{sec4.1.1}

We use the Odometry version {\color{blue}\cite{wimbauer2021monorec}} of the KITTI dataset {\color{blue}\cite{geiger2012we}}, which consists of 13,666 samples in the training dataset and 8,634 samples in the testing dataset. The frames are cropped and resized to ${256 \times 512}$, and the depth range is set from 0 to 80 ${m}$. Furthermore, we utilized the estimated pose {\color{blue}\cite{yang2018deep}} and the depth ground truth {\color{blue}\cite{uhrig2017sparsity}} to supervise the final depth estimation, and we leveraged the mask ground truth {\color{blue}\cite{wimbauer2021monorec}} to supervise the training of the super dynamic scene mask prediction module.

\subsubsection{Training}
\label{sec4.1.2}

All experiments are conducted on a single NVIDIA GPU 3090. To ensure comprehensive coverage of the dynamic scene compensation module, we first train the super dynamic scene mask prediction module and then train the other components based on the predicted mask prior. When training the super dynamic scene mask prediction module, we set the batch size to 32 and train for 80 epochs. We employ the Adam optimizer with the oncyclelr learning strategy {\color{blue}\cite{smith2019super}}, where the maximum learning rate is set to 2e-3 and the start and end division factors are both 10. It is important to note that, since dynamic scenes have a low pixel proportion and may not appear in some instances, during training, the IoU initially increases and then rapidly decreases, indicating a pattern collapse. As described in {\color{blue} Sec. \ref{sec3.2.2}}, we do not require a precise shape prediction. Therefore, we select the best-trained model with the highest covering proportion (${CP}$) as {\color{blue} Equ. (\ref{eq5})}, which provides sufficient candidates for the dynamic scene compensation module. Additionally, the training settings of the other components are the same as in {\color{blue}\cite{li2023learning}}.

\begin{equation}
\label{eq5}
CP = \frac{{TopK\left( {{m_t}} \right) \cap {{\tilde m}_t}}}{{{{\tilde m}_t}}}
\end{equation}

\subsection{Comparison}
\label{sec4.2}

\begin{table*}[!ht]
\caption{The quantitative comparisons between different methods with the same frame size ${256 \times 512}$. BB and SV represent backbone and supervision. \textbf{Bold} and \underline{underline} highlights represent the best and second-best performance.}
\label{table1}
\centering
\begin{tabular}{@{}llllllllll@{}}
\toprule
\textbf{Overall Scenes}                               & BB     & SV   & Abs Rel             & Sq Rel              & RMSE                & RMSElog             & ${\delta < 1.25}$   & ${\delta< {1.25^2}}$ & ${\delta<{1.25^3}}$ \\ \midrule
Ours                                                  & Res-18 & Full & \textbf{0.040}      &  \textbf{0.135}     & \textbf{1.994}      & \textbf{0.068}      & \textbf{0.980}      & \textbf{0.996}       & \textbf{0.999} \\
Manydepth{\color{blue}\cite{watson2021temporal}}      & Res-18 & Self & 0.071               & 0.343               & 3.184               & 0.108               & 0.945               & {\underline{0.991}}  & \underline{0.998} \\
DynamicDepth{\color{blue}\cite{feng2022disentangling}}& Res-18 & Self & 0.068               & 0.296               & 3.067               & 0.106               & 0.945               & {\underline{0.991}}  & {\underline{0.998}} \\
MonoRec{\color{blue}\cite{wimbauer2021monorec}}       & Res-18 & Semi & 0.050               & 0.290               & 2.266               & 0.082               & 0.972               & {\underline{0.991}}  & 0.996 \\
DMDepth{\color{blue}\cite{li2023learning}}            & Res-18 & Full & {\underline{0.043}} & {\underline{0.151}} & {\underline{2.113}} & {\underline{0.073}} & {\underline{0.975}} & {\textbf{0.996}}     & {\textbf{0.999}} \\ \midrule
Ours                                                  & Eff-b5 & Full & \textbf{0.040}      & \textbf{0.109}      & \textbf{1.815}      & \textbf{0.064}      & \textbf{0.984}      & \textbf{0.998}       & \textbf{0.999} \\
DMDepth{\color{blue}\cite{li2023learning}}            & Eff-b5 & Full& \underline{0.046} & \underline{0.155} & \underline{2.112} & \underline{0.076} & \underline{0.973} & \underline{0.996} & \textbf{0.999} \\
MaGNet{\color{blue}\cite{bae2022multi}}               & Eff-b5 & Full & 0.057 & 0.215 & 2.597 & 0.088 & 0.967 & \underline{0.996} & \textbf{0.999} \\ \midrule
 &  &  &  &  &  &  &  &  &  \\ \midrule

\textbf{Dynamic Scenes} & BB & SV & Abs Rel & Sq Rel & RMSE & RMSElog & ${\delta    < 1.25}$ & ${\delta    < {1.25^2}}$ & ${\delta    < {1.25^3}}$ \\ \midrule
Ours & Res-18 & Full & {\underline{0.163}} & {\underline{1.484}} & {\underline{5.309}} & {\underline{0.188}} & {\underline{0.791}} & {\underline{0.930}} & {\underline{0.974}} \\
Manydepth{\color{blue}\cite{watson2021temporal}} & Res-18 & Self & 0.222 & 3.390 & 7.921 & 0.237 & 0.676 & 0.902 & 0.964 \\
DynamicDepth{\color{blue}\cite{feng2022disentangling}} & Res-18 & Self & 0.208 & 2.757 & 7.362 & 0.227 & 0.682 & 0.911 & 0.971 \\
MonoRec{\color{blue}\cite{wimbauer2021monorec}} & Res-18 & Semi & 0.360 & 9.083 & 10.963 & 0.346 & 0.590 & 0.882 & 0.780 \\
DMDepth{\color{blue}\cite{li2023learning}} & Res-18 & Full & \textbf{0.118} & \textbf{0.835} & \textbf{4.297} & \textbf{0.146} & \textbf{0.871} & \textbf{0.975} & \textbf{0.990} \\ \midrule
Ours & Eff-b5 & Full & \underline{0.135} & \underline{1.002} & \underline{4.774} & \underline{0.160} & 0.829 & \underline{0.968} & \underline{0.992} \\
DMDepth{\color{blue}\cite{li2023learning}} & Eff-b5 & Full& \textbf{0.111} & \textbf{0.768} & \textbf{4.117} & \textbf{0.135} & \textbf{0.881} & \textbf{0.980} & \textbf{0.994} \\
MaGNet{\color{blue}\cite{bae2022multi}} & Eff-b5 & Full & 0.141 & 1.219 & 4.877 & 0.168 & \underline{0.830} & 0.955 & 0.986 \\ \bottomrule
\end{tabular}
\end{table*}

\begin{figure*}[!ht]
\centering
\includegraphics[scale=.188]{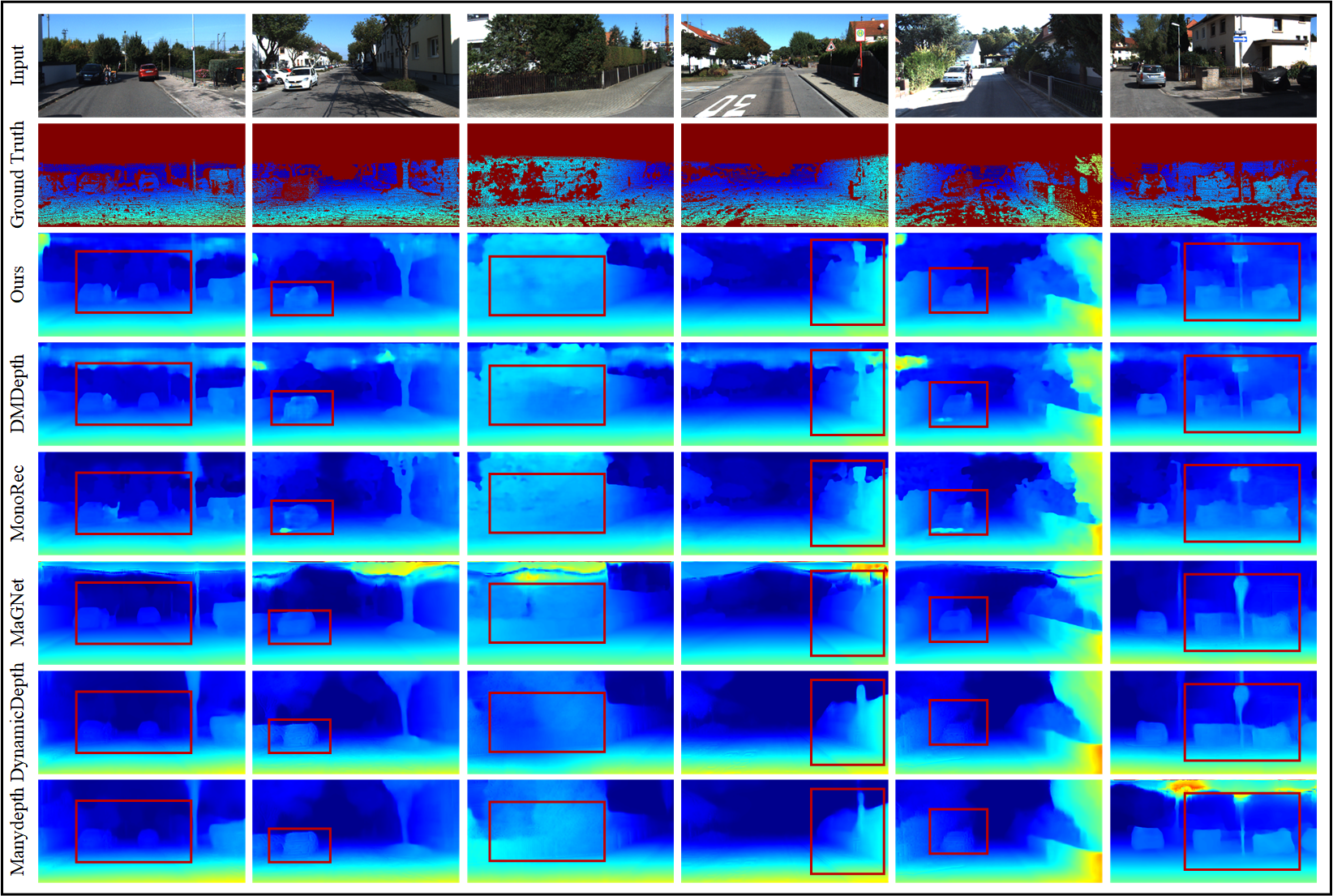}
\caption{The qualitative comparisons between different methods on the KITTI dataset.}
\label{fig4}
\end{figure*}

We compared our method with Manydepth {\color{blue}\cite{watson2021temporal}}, DynamicDepth {\color{blue}\cite{feng2022disentangling}}, MonoRec {\color{blue}\cite{wimbauer2021monorec}}, MaGNet {\color{blue}\cite{bae2022multi}}, and DMDepth {\color{blue}\cite{li2023learning}}. The quantitative and qualitative results are shown in {\color{blue}Table \ref{table1}} and {\color{blue}Fig. \ref{fig4}}. In overall scenes, our method outperforms all other methods in terms of all evaluated metrics. Compared to second best method DMDepth, our method decreases Abs Rel, Sq Rel, RMSE by 6.98\%, 10.57\%, 5.63\%, respectively. In dynamic scenes, our method ranks second, following DMDepth, but still significantly outperforms the other methods. Moreover, the cue fusion within our method exhibits a 3.24 ${G}$ FLOPs and achieves a 227.41 FPS, surpassing the performance of DMDepth (3.82 ${G}$ FLOPs and 171.73 FPS). A detailed efficiency comparison will be further discussed in {\color{blue} Sec. \ref{sec4.3}}.

\subsection{Ablation Study}
\label{sec4.3}

\subsubsection{Attention in Cue Fusion}
\label{sec4.3.1}

To assess the efficiency and precision of various attention mechanisms used in cue fusion, we conducted four comparison groups. For a fair comparison, the FLOPs of our case only contain the deformable attention module and the dynamic scene compensation module. The second group (Deform) employed sparse deformable attention alone {\color{blue}\cite{zhu2021deformable}} with a downsampling factor of ${2 \times }$. The third group (Local) utilized local window attention {\color{blue}\cite{liu2021swin}} for overall patches with a downsampling factor of ${4 \times }$. The fourth group (full) employed full attention as described in {\color{blue}\cite{li2023learning}} with a downsampling factor of ${4 \times }$. The fifth group (Swin+Local) replaces the deformable attention module as a one-stage Swin transformer with a downsampling factor of ${2 \times }$, and employs the remaining two modules to compensate for dynamic scenes. In {\color{blue}Table \ref{table2}}, when considering overall scenes, deformable attention enhances precision as it refines cue granularity, while maintaining high cue fusion efficiency. Regarding dynamic scenes, deformable attention exhibits a significant decrease in precision compared to full attention due to its sparse relationship learning, as discussed in {\color{blue} Sec. \ref{sec3.2.1}}. In addition, the superior performance of local window attention compared to deformable attention demonstrates that local dense relationship learning enhances spatial exploration among different cues. The fifth group enhances the local relationship learning between cues at the fine granularity, while omitting global relationship exploration. This combination of attention mechanisms contributes to improved performance specifically in dynamic scenes, as opposed to overall scenes. It is worth noting that that deploying a one-stage Swin transformer on fine granularity results in a significant increase in FLOPs, reaching 7.21 ${G}$, thereby contradicting the high-efficiency objective. In contrast, we utilize a super dynamic scene mask to mediate between deformable and local window attentions, adapting them to different scenes and achieving a trade-off of precision between overall and dynamic scenes. Nonetheless, the computational cost incurred by attention in our cue fusion is a mere 2.84 ${G}$ FLOPs. Even with the incorporation of super dynamic scene mask prediction, the total cost for our cue fusion can remain at 3.24 ${G}$ FLOPs, resulting in a 15.2\% cost reduction compared to full attention.

\begin{table}[h]
\caption{The application of different types of attention in cue fusion. DS represents the downsampling factor. {\color{cyan} Cyan} and {\color{purple} Purple} highlights denote the overall and dynamic scenes performances, respectively. \textbf{Bold} and \underline{underline} highlights represent the best and second-best performance.}
\label{table2}
\centering
\resizebox{\linewidth}{!}{
\begin{tabular}{@{}llllllll@{}}
\toprule
             & DS             & {\color{cyan}Abs Rel} & {\color{cyan}${\delta  < 1.25}$}  & {\color{purple}Abs Rel} & {\color{purple}${\delta  < 1.25}$} & FLOPs/${G}$      & FPS \\ \midrule
Ours         & /              & \underline{0.040}     & \underline{0.980}                 & 0.163                   & 0.791                              & 2.84             & 305.51 \\
Deform       & ${2 \times }$  & \textbf{0.039}        & \textbf{0.983}                    & 0.265                   & 0.602                              & \textbf{1.03}    & \textbf{834.46} \\
Local        & ${4 \times }$  & 0.079                 & 0.919                             & 0.200                   & 0.694                              & \underline{1.69} & \underline{510.11} \\
Full         & ${4 \times }$  & 0.043                 & 0.975                             & \textbf{0.118}          & \textbf{0.871}                     & 3.82             & 171.73 \\
Swin+Local   & /              & 0.045                 & 0.975                             & \underline{0.157}       & \underline{0.798}                  & 7.21             & 131.76    \\ \bottomrule
\end{tabular}
}
\end{table}

\subsubsection{Lightweight of Mask Prediction}
\label{sec4.3.2}
As GSDC Transformer introduces a super dynamic scene mask prediction module to guide the compensation, in this section, we describe its lightweight design without compromising fusion efficiency. Since the ${TopK( \cdot )}$ operation in {\color{blue} Equ. (\ref{eq2})} corresponds to the one shown in {\color{blue}Fig. \ref{fig2}b}, we evaluate ${CP}$ metric for ${{N_\vartheta } = 6}$ and ${{N_\vartheta } = 12}$. As shown in {\color{blue} Table \ref{table3}}, row 2 represents no downsampling and upsampling in the prediction of ${{{\cal M}_t}}$, Despite the relatively low shape precision (IoU), it still enables ${{m_t}}$ to provide candidates to the dynamic scene compensation module. Since the downsampling scale of the dynamic compensation module is ${4 \times }$, we explore a lightweight prediction approach in row 3 and 4, which directly performs downsampling and upsampling during the prediction of ${{{\cal M}_t}}$. Experimental results indicate that this coarse-scale prediction does not cause a significant drop in ${CP}$. Moreover, the FLOPs for row 3 amount to only 6.3\% of those in row 2, while the FLOPs for row 4 are reduced to just 0.40 ${G}$. Another alternative method in row 5 employs a MaxPooling layer with a kernel size of ${\kappa  \times \kappa }$ at the prediction head, and it also achieves good performance.

\begin{table}[h]
\small
\caption{The lightweight design of the super dynamic scene mask prediction module. DS, US, and MP represent downsampling, upsampling, and MaxPooling.}
\label{table3}
\centering
\begin{tabular}{@{}lllll@{}}
\toprule
 & IoU & ${C{P_{{N_\vartheta } = 6}}}$ & ${C{P_{{N_\vartheta } = 12}}}$ & FLOPs/${G}$ \\ \midrule
No DS \& US & 0.065 & 0.946 & 0.970 & 25.37 \\
DS \& US ${\left( {4 \times   } \right)}$ & 0.053 & 0.944 & 0.971 & 1.59 \\
DS \& US ${\left( {8 \times   } \right)}$ & 0.046 & 0.936 & 0.963 & 0.40 \\
DS \& MP ${\left( {4 \times   } \right)}$ & 0.049 & 0.939 & 0.967 & 1.59 \\ \bottomrule
\end{tabular}

\end{table}

\subsubsection{Parameter Setting}
\label{sec4.3.3}

For efficient fusion, we set the default values of ${{N_s}{\rm{ = }}4}$ and ${{N_\vartheta }{\rm{ = }}12}$. Upon counting all samples in the dataset {\color{blue}\cite{geiger2012we}}, we found that within the range of ${\mu  \pm 2\sigma }$ in the normal distribution, the proportion of super tokens attribute to dynamic scenes to all super tokens is approximately 4\%. However, with 6 and 12 candidates in our setting, the proportions increase to 5\% and 10\%, respectively, indicating that they cover almost all dynamic scenes in all samples. Consequently, As shown in {\color{blue} Table \ref{table4}}, row 2 and row 3 exhibit similar performances, and further increasing ${{N_\vartheta }}$ to 25 does not lead to a further enhancement. Furthermore, increasing ${{N_s}}$ to 8 results in improved overall performances at the expense of computational cost.

\begin{table}[ht]
\caption{The quantitative comparison under different parameter settings. {\color{cyan} Cyan} and {\color{purple} Purple} highlights denote the overall and dynamic scenes performances, respectively.}
\label{table4}
\centering
\resizebox{\linewidth}{!}{
\begin{tabular}{@{}llllllll@{}}
\toprule
${{N_s}}$ & ${{N_\vartheta }}$ & {\color{cyan}Abs Rel} & {\color{cyan}${\delta  < 1.25}$} & {\color{purple}Abs Rel} & {\color{purple}${\delta  < 1.25}$}  & FLOPs/${G}$ & FPS \\ \midrule
4 & 12 &0.040 &0.980 & 0.163 & 0.791 & 2.84 & 305.51 \\
4 & 6 & 0.039 & 0.981 & 0.166 & 0.768 & 2.84 & 313.39 \\
4 & 25 & 0.038 & 0.983 & 0.170 & 0.772 & 2.84 & 299.41 \\
8 & 12 & 0.037 & 0.983 & 0.175 & 0.772 & 2.89 & 279.16 \\ \bottomrule
\end{tabular}
}
\end{table}

\section{Conclusion}
This paper proposes the GSDC Transformer as an efficient and effective component for cue fusion in monocular multi-frame depth estimation. The primary objective of this paper is to introduce a design paradigm about granularity and sparseness in cue fusion, considering both overall and dynamic scenes. We enhance the granularity of cue representation and utilize deformable attention to learn their relationships at a fine scale, resulting in a significant improvement in overall precision. However, due to the significant displacement of moving objects between frames, sparse attention is inadequate for exploring the relationships between their cues, resulting in a precision drop. To address this issue, we first represent scene attributes in the form of super tokens without relying on precise shapes. Subsequently, we select a specific number of candidates using the TopK function. Within each super token attribute to dynamic scenes, the relevant cues are gathered, and local dense relationships are learned to enhance cue fusion. Experiments demonstrate that our method achieves state-of-the-art performance on the KITTI dataset, ranking first and second for overall and dynamic scenes, respectively, and our fusion method achieves nearly a 20\% reduction in FLOPs.

While our method demonstrates competitive performance, its precision in dynamic scenes falls short of certain state-of-the-art methods. In future work, we will continue to refine the compensation mechanism and framework to bolster performance specifically in dynamic scenes. Notably, the training of the super dynamic scene mask prediction module needs ground truth for dynamic scenes, incurring additional costs for manual labeling. Consequently, we also aim to devise an adaptive compensation to mitigate the reliance on this manual labeling.

\bibliographystyle{IEEEtran}
\bibliography{refs}{}

\end{document}